\documentclass[11pt,a4paper]{article}
\usepackage[hyperref]{acl2020}
\usepackage{times}
\usepackage{latexsym}

\usepackage{amsmath}
\usepackage[linesnumbered,ruled]{algorithm2e}
\usepackage{tablefootnote}

\usepackage{microtype}

\aclfinalcopy % Uncomment this line for the final submission
\title{A Complex KBQA System using Multiple Reasoning Paths}

\author{Kechen Qin$^{1}$ ~~~~ Yu Wang$^{2}$ ~~~~ Cheng Li$^{1}$ ~~~~ Kalpa Gunaratna$^{2}$\\ ~~~~ \textbf{Hongxia Jin$^{2}$} ~~~~ \textbf{Virgil Pavlu$^{1}$} ~~~~ \textbf{Javed A. Aslam$^{1}$}\\
  $^{1}$Khoury College of Computer Sciences,
  Northeastern University\\
  $^{2}$Samsung Research America\\
{$^{1}$\tt qin.ke@husky.neu.edu} ~~~~ {\tt \{chengli,vip,jaa\}@ccs.neu.edu}\\
{$^{2}$\tt \{yu.wang1,kaushik.g1,hongxia.jin\}@samsung.com }
   \\}

\date{}

\usepackage{tikz}

\begin{document}

\maketitle

\begin{abstract}
 Multi-hop knowledge based question answering (KBQA) is a complex task for natural language understanding. Many KBQA approaches have been proposed in recent years, and most of them are trained based on labeled reasoning path. This hinders the system's performance as many correct reasoning paths are not labeled as ground truth, and thus they cannot be learned. In this paper, we introduce an end-to-end KBQA system which can leverage multiple reasoning paths' information and only requires labeled answer as supervision. We conduct experiments on several benchmark datasets containing both single-hop simple questions as well as muti-hop complex questions, including WebQuestionSP (WQSP), ComplexWebQuestion-1.1 (CWQ), and PathQuestion-Large (PQL), and demonstrate strong performance.
\end{abstract}

\begin{figure}[t]
 \centering
 \includegraphics[width=1\linewidth]{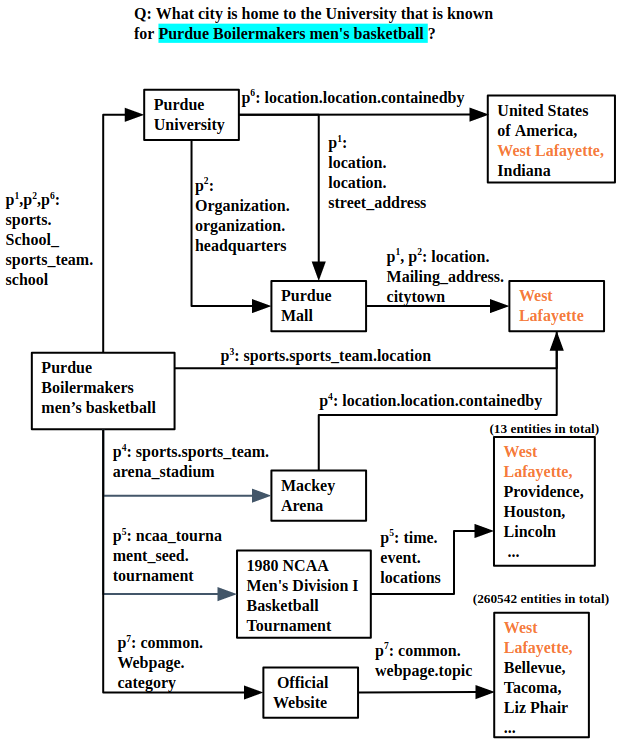}
 \caption{One QA example with Multiple Reasoning Paths from \textsc{ComplexWebQuestion}-1.1. The blue color highlighted is the extracted topic entity. Each square represents an entity, and the arrows represent the relations. Reasoning path $\mathbf{p}^1$ to $\mathbf{p}^4$ are the correct ones containing meaningful reasoning paths to the final answer. $\mathbf{p}^5$ and $\mathbf{p}^6$ are the ``second choice'' paths that generate a larger final answer set containing some wrong entities. $\mathbf{p}^7$ is the wrong one as its reasoning path is totally not interpretable and the answer set is huge.}
 \label{QAPaths}
\end{figure}

\section{Introduction}

Knowledge-based question answering (KBQA) is the task of finding answers to questions by processing a structured knowledge base $\mathcal{KB}$. %where the beliefs are stored as triples containing two entities and the relation linking them. 
A $\mathcal{KB}$ consists of a set of entities $\mathcal{E}$, a set of relations $\mathcal{R}$, and a set of literals $\mathcal{S}$. A knowledge base fact is defined as $(h,r,t)$, where $h\in \mathcal{E}$ is the head entity, $t \in \mathcal{E} \bigcup \mathcal{S}$ is the tail entity/literal, and $r\in \mathcal{R}$ is the directed relation between $h$ and $t$. To answer a simple single-relation question (\emph{i.e.} a 1-hop question) such as: \textit{``Who is the president of the United States?''}, %a KBQA system first identifies the topic/focus entity (\emph{i.e.} United States) and the relation (\emph{i.e.} ``president '') asked in the question, then searches for the entity by matching the entity-relation tuple $\textless$United States, president, $?\textgreater$ over KB. \kalpa{use of a topic entity is approach specific. Many KBQA systems exist that do not use topic entity. E.g., template based QA}
a typical KBQA system first identifies the entity (\emph{i.e.} United States) and the relation (\emph{i.e.} ``president'') asked in the question, and then searches for the answer entity by matching the entity-relation tuple $\textless$United States, president, $?\textgreater$ over $\mathcal{KB}$.

While a single-hop question can be answered by searching a predicate relation in $\mathcal{KB}$, it is much harder to answer more complex multi-hop questions containing multiple entities and relations with constraints. For instance, for complex compositional questions, it is not easy to extract all the relations correctly together with their head and tail entities in the right order. For complex conjunction questions that requires a conjunction of multiple evidences, it is even more difficult to correctly extract all the reasoning paths included.

Most prior works on multi-hop KBQA focus on learning a single given ground truth reasoning path for each question, and outputting the most possible reasoning path during prediction \cite{DBLP:conf/coling/ZhouHZ18,DBLP:journals/corr/abs-1801-09893,DBLP:conf/adbis/YuHYZW18,DBLP:conf/ijcai/LanW019}. However, it is common that $\mathcal{KB}$ has many alternative paths leading to the correct answer, of various reasoning qualities. These alternative reasoning paths are usually not provided as ground truth by the human annotators. %It is hard to pre-define the maximum number of hops in a complex question because different questions may need different number of hops to reach their answers.
For example, Figure \ref{QAPaths} shows 7 reasoning paths $\mathbf{p}^n={e^n_0\rightarrow r^n_1 \rightarrow e^n_1 \rightarrow \cdots \rightarrow e^n_{ans}}\ (n=\lbrace 1, \dots, 7 \rbrace)$ leading to an answer set containing the correct answer \textit{``West Lafayette''} for a given question \textit{``What city is home to the University that is known for Purdue Boilermakers men's basketball?''}%\footnote{An exchangeable way to define relation path is omitting entities in the definition, because entity is determined by given the topic entity, relations, and the knowledge base. For simplicity, we simplify this relation path definition as a list of sequential relations $(r_{1}^n, r_{2}^n, \cdots, r_{T}^n)$ in the following part of this paper.}
, but only the reasoning path $\mathbf{p}^1$ is labeled as the correct path in the dataset. A model trained with only $\mathbf{p}^1$ as supervision is likely to miss other paths which are also valid. For example, it will probably map a similar question \textit{``What city is home to the stadium that is known for Los Angeles Lakers?''} to path $\mathbf{p}^1$, but fail to associate it with $\mathbf{p}^3$ or $\mathbf{p}^4$, because $\mathbf{p}^3$ or $\mathbf{p}^4$ contain different types of relations. However, $\mathbf{p}^1$ is a wrong reasoning path for that test question.

As the example shown in Figure \ref{QAPaths}, there are four paths ($\mathbf{p}^1$,$\mathbf{p}^2$,$\mathbf{p}^3$,$\mathbf{p}^4$) pointing to the exact answer set containing only the answer entity, and thus can be treated as ground truth paths when training. Comparatively, reasoning paths $\mathbf{p}^5$ and $\mathbf{p}^6$ lead to a larger final entity set containing the correct answer \textit{``West Lafayette''} but also other entities. These two paths can be considered as inferior to the top 4 paths; however, it is still worth including them in the training as a ``second choice'', as it is not difficult to extract the correct answer from final sets by additional post-processing. For example, a simple filter can be applied to filter out \textit{``United States of America''} and \textit{``Indiana''} from the predicted set, as they are not cities. Path $\mathbf{p}^7$ is bad because it is not interpretable, in addition to the final answer set being exaggeratedly large with invalid answers. Hence, path $\mathbf{p}^7$ should not be considered as a training path for this question. Unfortunately, it is not possible for any existing models to use multiple good/inferior paths, but not the bad ones, since current models are only trained with a single path for each question answer pair.

%Although someone may claim that we can use multiple relation paths as ground truths for one question to improve the coverage and performance, it is still not feasible to label all valid relation paths considering the complexity of ground truth label preparation. 
%\kechen{combine this paragraph with the next paragraph? now we have two separate paragraphs to first introduce the issue and then use an example to explain it. should we combine them together?}

In this paper, we propose an end-to-end multi-hop KBQA system, which can leverage the training information from multiple reasoning paths without using any path annotations. We model the reasoning path as a latent variable, and propose supporting training and prediction methods. The system can output diverse reasoning paths, and reward the ``better'' paths over the inferior ones by assigning ``better'' paths higher probabilities. Our method can be applied to most KBQA systems to predict the answer, and can be used with any model architecture. We achieve strong performance on three popular KBQA datasets. Experimental results show that our model performs especially well on multi-hop question, and in particular on complex questions that cannot be solved with a single reasoning path.

Our method does not need training paths annotation (only the question, and head and final entities), since it can sample the paths from the $\mathcal{KB}$ graph. This is of enormous pratical importance, because in practice questions and answers are easy to collect (sometimes for free), but path annotation is very labor-intensive and expensive.

\begin{figure*}[t]\centering
\includegraphics[width=2.0\columnwidth]{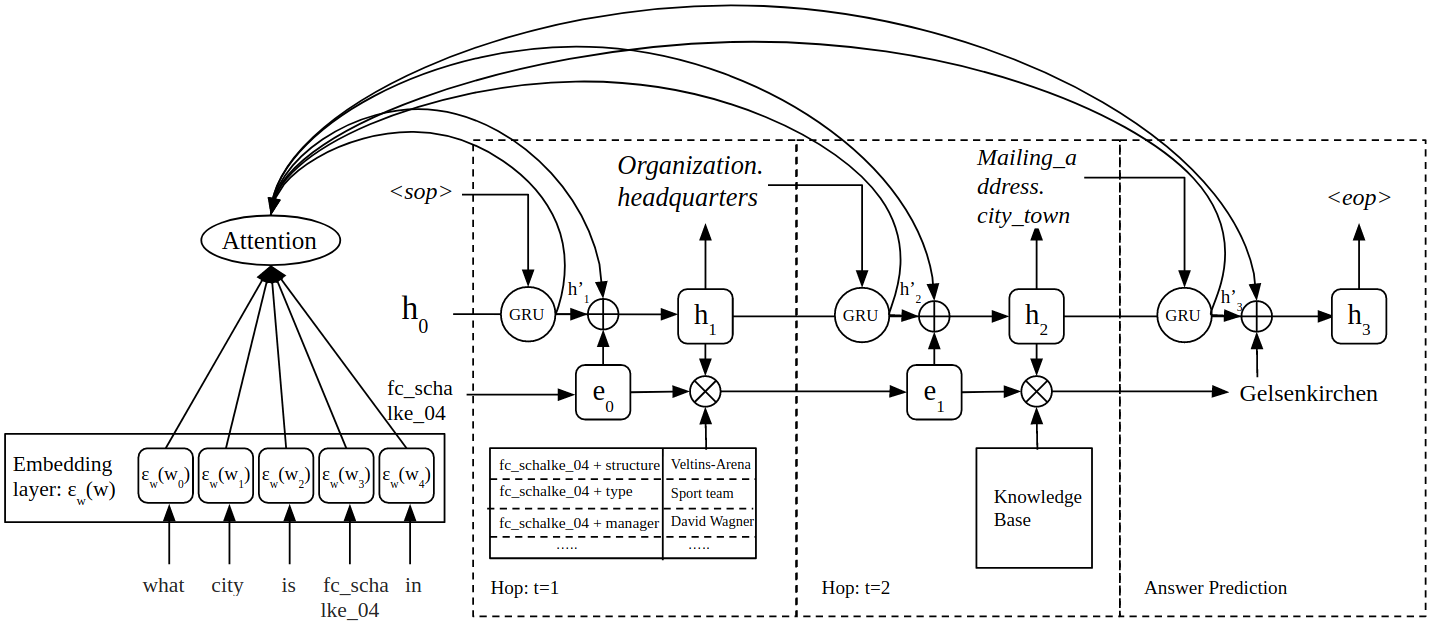}
\caption{\fontsize{10}{12}\selectfont An illustration of how our model works with a QA pair \textit{"What city is fc\_schalke\_04 in?"} and \textit{Gelesenkirchen}. The entity linker extracts \textit{fc\_schalke\_04} as the topic entity. We only show one possible paths here: $r_1$ is \textit{Organization.headquarters} and $r_2$ is \textit{Mailing\_address.city\_town}, our model can be used to output the probability of this given path. The symbol $\bigoplus$ represents concatenation, and $\bigotimes$ represents knowledge base lookup. }
\label{fig:model}
\end{figure*}

\section{Model}
We first introduce some notations. For a given question $q$ and its topic entity $e_0$ (identified by entity linking tool), a reasoning path is a sequence in the form $\mathbf{p} = (e_0, r_{1},e_{1},r_{2}, \cdots,e_{T-1}, r_{T})$ that points to the answer entity $e_T=y$. That is, $\mathbf{p}\rightarrow e_T=y$. Each step $(e_{t-1},r_t,e_t)$ is a valid fact in the knowledge base $\mathcal{KB}$. Our goal is to build a model that can use multiple paths $\mathbf{p}$ to predict answer $y$ given question $q$ and topic entity $e_0$. In this section, we first present the design of our model architecture, and then explain the training and inference algorithms in detail. 

\subsection{Model Architecture}

%Figure \ref{fig:model} illustrates the architecture of our model. All the words $w_0,w_1,\cdots,w_{|q|-1}$ in the given question $q$ are first sent to a fixed embedding layer to acquire word embeddings $\varepsilon_w(w_0),\varepsilon_w(w_1),\cdots,\varepsilon_w(w_{|q|-1})$. \cheng{Does the symbol match the one in diagram?} To reduce computational cost, the embedding layer is pre-trained and not updated during training. %The word embeddings are combined in different ways based on attention weights to show different reasoning focus at each hop.
% $\varepsilon(\cdot)$ function represents gathering embedding of the input from the corresponding pre-trained embeddings matrix. 

%\subsubsection{Reasoning Decoding} The decoding module consists of relation prediction module and entity prediction module.

Figure \ref{fig:model} illustrates the architecture of our model. We model path probabilities using recurrent neural network with gated recurrent units (GRU). At a timestep $t$, the input hidden representations of GRU unit and predicted relation are denoted by $h_{t-1}$ and $r_t$ respectively. The model relies on the attention mechanism~\cite{DBLP:journals/corr/BahdanauCB14} to produce a question context vector $c_t$. Specifically, all the words $w_0,w_1,\cdots,w_{|q|-1}$ in the given question $q$ are first sent to a fixed embedding layer to acquire word embeddings $\varepsilon_w(w_0),\varepsilon_w(w_1),\cdots,\varepsilon_w(w_{|q|-1})$. Next we apply GRU to produce a temporary hidden state $h_{t}'=GRU(h_{t-1}, r_{t-1})$, and then apply a parameterized feed-forward neural network $a$ to calculate the similarity score $u_{tk} = a(h'_{t},\varepsilon_w(w_k))$ of two inputs $h'_{t}$ and $\varepsilon_w(w_k)$, and then these scores are normalized into attention weights $\alpha_{tk}=\frac{\exp (u_{tk})}{\sum_{0\leq j\leq |q|-1}\exp (u_{tj})}$, which are used to produce the question context vector $c_t=\sum_{0\leq j\leq |q|-1}\alpha_{tj}\varepsilon_w(w_j)$. In this fashion, word embeddings are combined in different ways based on attention weights to show different reasoning focuses at each timestep.
% \begin{align}
% h'_{t} = GRU(h_{t-1}, r_{t-1}) 
% \end{align}
% \vspace{-3ex}
% \begin{align}
% u_{tk} = a(h'_{t},\varepsilon(w_k))
% \end{align}
% \vspace{-3ex}
% \begin{align}
% \alpha_{tk} = \frac{\exp (u_{tk})}{\sum_{0\leq j\leq |q|-1}\exp (u_{tj})}
% \end{align}
% \vspace{-1ex}
% %\[\alpha_{tk} = \frac{u_{tk}}{\sum_{j}u_{tj}}\]
% \begin{align}
% c_t = \sum_{0\leq j\leq |q|-1}\alpha_{tj}\varepsilon(w_j)
% \end{align}

 %After having all above computations done, 

The model then concatenates temporary hidden state $h_{t}'$, entity representation $\varepsilon_e(e_{t-1})$, and question context $c_t$ together, and passes the concatenation through a linear transformation $f$ with ReLU activation to obtain the hidden state $h_t=ReLU(f([h'_{t}; \varepsilon_e(e_{t-1}); c_t]))$. This process is recurrently done until the model predicts a stop symbol \textit{$\textless$eop$\textgreater$}\footnote{This stop mechanism is the same as how it works in a vanilla RNN. Similarly, we also attach $\textless$sop$\textgreater$ to the beginning of each sequence to denote the start state. We will omit these symbols in formulas for simplicity.}. Note that the vanilla RNN attention model only has $h_{t}'$ and $c_t$ when calculates $h_t$. We add entity representation into the calculation, since entity captures important information in the reasoning path. 

\subsection{Probabilities and Objective Function} 
The probability of predicting the $k$-th relation $\gamma_k$ in $\mathcal{R}$ at timestep $t$ is:
\begin{align*}
&p(r_t=\gamma_k|q,e_0,r_1,\cdots,e_{t-1})\\
 &= \frac{\exp <h_t,\varepsilon_r(\gamma_k)>}{\sum_j\exp <h_t,\varepsilon_r(\gamma_j)>}
\end{align*}
where $\varepsilon_r$ is the embedding function, $<>$ is the dot product between two inputs.

Given the previous entity $e_{t-1}$ and relation $r_t$, the next matched entity may not be unique when we query the knowledge base. %(another way to collect this entity is via a soft lookup with a key-value memory network structure. We provide more details in the appendix.) 
For example, if $e_{t-1}$=``united states'', and $r_t=$ ``president of'', then the resulting entity has 45 possibilities. Since we do not have additional constraints, all of them are equally likely to be selected, and hence we define:
\begin{align}
  %&p(e_t|q,e_0,r_1,\cdots,e_{t-1},r_t)\\\nonumber
  %=&\;
  &p(e_t|e_{t-1},r_t)\\\nonumber
  =&\begin{cases}
  1/M & \text{if }e_t\text{ is one of the }M\text{ matched entities} \\\nonumber
  0 & \text{if }e_t\text{ is not a matched entity}
\end{cases} 
\label{eq:equal}
\end{align}

Thus the probability of a path containing both entities and relations can be computed using the chain rule:

\begin{align}
&p(\mathbf{p}|q)\nonumber\\
=& \prod_{t=1}^{T-1}p(e_t|e_{t-1},r_t)\prod_{t=1}^{T}p(r_t|q,e_0,r_1,\cdots,e_{t-1}) 
\end{align}

We assume that there are multiple valid paths $\mathbf{p}\in \mathcal{P}$ that can lead to the correct answer $y$ and they are not given by the annotator in the dataset. We treat these paths as hidden variables and we marginalize them out to compute the probability of getting the answer $y$:

\begin{align}
&p(y|q)\nonumber\\
=&\sum_{\mathbf{p}\in\mathcal{P}} [p(e_{T(\mathbf{p})}=y|\mathbf{p},q)p(\mathbf{p}|q)] \nonumber\\
=&\sum_{\mathbf{p}\in\mathcal{P}}\prod_{t=1}^{T(\mathbf{p})} [p(e_t|e_{t-1},r_t) p(r_t|q,e_0,r_1,\cdots,e_{t-1})]
\label{eq:marginal}
\end{align}
where $\mathcal{P}$ is a set of all valid paths leading to the answer $y$, and $T(\mathbf{p})$ is the number of hops in the path $\mathbf{p}$.

To train our model, we would like to maximize the answer probability $p(y|q)$ using only the given answer for each training instance. To make prediction on each test case, we would like to find the answer $y$ with the highest probability.

It is a novel way that we define answer probability as in (\ref{eq:marginal}) in the KBQA task. Most of the existing methods assume the availability of a single ground truth path annotation and aim to maximize the probability of the given path~\cite{DBLP:conf/coling/ZhouHZ18}. As we will demonstrate later in the Section \ref{sec:cop}, considering multiple paths leads to a better model performance.

%To further improve the diversity of the outputs, we follow \newcite{}'s idea to optimize mutual information instead of log likelihood. Specifically, we observe that 

\subsection{Training}

In order to train our model by maximizing the marginalized answer probability given in (\ref{eq:marginal}), it requires summing over all valid reasoning paths from the topic entity to the answer entity in knowledge base. Thus computing this objective exactly can be intractable. As shown in the early example, some reasoning paths ($R_5, R_6, R_7$ in Figure \ref{QAPaths}) are not very helpful for training, thus should be either removed from training or assigned low probabilities. To achieve this goal, we first apply depth first search (DFS) algorithm with maximum 3 hops to get valid path candidates. The algorithm starts the traversal from the topic entity node, and ends at the answer entity node. All possible paths between the topic entity and the answer entity within 3 hops are extracted as candidates. We then set a threshold to remove paths which point to too many entities at the last hop. To further filter out bad reasoning paths, we propose to dynamically choose reasoning paths deemed as most probable by the current model during training. The overall training procedure is summarized in Algorithm \ref{alg:train}. Note that training with this algorithm does not require ground truth reasoning path label. Labeled reasoning path is a plus, but not necessary. If it is given, we can either include the ground truth paths in $\mathcal{P}$, or use them to initialize model training.

\begin{algorithm}
 \SetKwInOut{Input}{Input}
 \SetKwInOut{Output}{Output}

 % \underline{function Euclid} $(a,b)$\;
 \Input{KBQA dataset $(q^{n},y^{n}, e_0^{n}),n=1,2,\cdots,N$, \\
 Knowledge Base $\mathcal{KB}$, \\
 Threshold $k_1$ and $k_2$. }
 \Output{Trained model parameters}
 \ForEach{instance $(q^{n},y^{n}, e_0^{n})$}{Use DFS algorithm to get a set of paths $\mathcal{P}^n$ from $e_0^{n}$ to $y^{n}$.\\
 Remove from $\mathcal{P}^n$ paths that point to more than $k_1$ entities.\\}
 
 % Initialize model parameters\\
 \ForEach {batch}{
 % \ForEach {batch }{
  \ForEach{$(q^{n},y^{n}, e_0^{n})$ in the batch}{
  Get top $k_2$ paths in $\mathcal{P}$ sorted by $p(\mathbf{p}|q)$ based on current model:
		 	$\tilde{\mathcal{P}}^n = \{{\mathbf{p}}^n_{1},\cdots,{\mathbf{p}}^n_{k_2} \}$\\
  } 
    Update model parameters by maximizing $\sum\limits_{(q^{n},y^{n}, e_0^{n}) \in \text{batch}} \log \sum\limits_{\mathbf{p} \in \tilde{\mathcal{P}}^n} p(y^{n}|\mathbf{p},q^{n}) P(\mathbf{p}|q^{n}) $
  }

 % }
 \caption{Our training method}
 \label{alg:train}
\end{algorithm}
\subsection{Prediction} \label{sec:pmi}

During the prediction, we aim to select the answer $y$ with the highest marginalized probability $p(y|q)$ as defined in (\ref{eq:marginal}). Similar to training, we need to approximate the sum with selected paths from $\mathcal{P}$. We use a modified beam search to find paths that have high probabilities. We add two constraints to standard beam search to only select the valid paths that match the knowledge base: (1) The first relation $r_1$ should connect to the topic entity $e_0$. (2) Each triple $(e_{t-1},r_t,e_t)$ should match a fact in KB. Given the set of paths collected as above, we can then collect a set of candidate answers that these paths point to. For each answer $y$, we evaluate its probability $p(y|q)$
approximately using the collected paths, and among them we output the answer with the highest probability.

%Given a set of paths collected, we can then obtain a candidate answer set $Y$ which these paths point to. For each answer $y \in Y$, we evaluate its probability $p(y|q)$ using the collected reasoning paths. Finally, among all $y \in Y$ we output the answer $y^*$ with the highest $p(y|q)$, \emph{i.e.} $y^*=\argmax_{y \in Y}p(y|q)$.

Additionally, we observe that it could be beneficial to de-emphasize the impact of the topic entity during prediction, as noted in \cite{DBLP:conf/naacl/LiGBGD16}, which can improve inference performance by avoiding generating generic predictions and reducing overfitting. Specifically, instead of searching $y^*$ that maximizes $p(y|q)$, we can find an answer that maximizes $\dfrac{p(y|q)}{p(y|e_0)}$, where $p(y|e_0)$ is the probability of getting the answer $y$ when the question only contains the topic entity word. Mathematically, one can show that this is equivalent to maximizing the point-wise conditional mutual information PMI$((y;q\backslash e_0)|e_0)$ between $y$ and $q\backslash e_0$ given $e_0$, where $q\backslash e_0$ stands for the question with the topic entity term removed. Further discussion can be found in Section \ref{sec:case}.

% \begin{equation}
% \begin{aligned}
% &\log PMI(y;f^{-ent}|f^{ent})\\
% =& \log\frac{P(y,f^{-ent}|f^{ent})}{P(y|f^{ent})P(f^{-ent}|f^{ent})}\\
% =& \log\frac{P(y|f^{-ent},f^{ent})}{P(y|f^{ent})}\\
% =& \log P(y|f^{-ent},f^{ent})-\log P(y|f^{ent})\\
% =& \log\sum_{\mathbf{p}\in\mathcal{P}}P(y,\mathbf{p}|f^{-ent},f^{ent}) - \log\sum_{\mathbf{p}\in\mathcal{P}}P(y,\mathbf{p}|f^{ent})\\
% \end{aligned}
% \label{eq:pmi}
% \end{equation}

% where $f^{-ent}$ represent all info without using entity\kechen{unify $P(x),p(x)$}

% \kechen{mutual info}

\section{Results and Analysis}
\subsection{Experimental Setup}
We conduct experiments on 3 multi-hop KBQA datasets, \textsc{WebQuestionSP} (WQSP) \cite{DBLP:conf/acl/YihCHG15}, \textsc{ComplexWebQuestion}-1.1 (CWQ) \cite{DBLP:journals/corr/abs-1807-09623}, and \textsc{PathQuestion-Large} (PQL) \cite{DBLP:conf/coling/ZhouHZ18}, and use the original train/dev/test split. WQSP is a dataset that has been widely used for relation extraction and end-to-end KBQA tasks, which contains 1 or 2 hops questions. CWQ dataset is designed to study complex questions by adding more constraints to questions in \textsc{WebQuestionSP}. PQL is a small dataset used to study sequential questions. Its original release contains two subsets: PQL2H and PQL3H, which contains only 2-hop and 3-hop questions correspondingly. \newcite{DBLP:conf/naacl/ChenCCNK19} then combined these two subsets and renamed the unified dataset as PQL+. All of the three datasets use Freebase \cite{freebase:datadumps} as the supporting knowledge base. Table \ref{tab:stats} contains statistics of these datasets. 

\begin{table}[h]\centering
\resizebox{0.8\columnwidth}{!}{
\begin{tabular}{|l|c|c|c|c|c|}
\hline
& \#train & \#valid & \#test & max\_hops & \textgreater 1 path\\
\hline
WQSP & 2677  & 297   & 1639  & 2 & 79.4\%   \\
CWQ  & 27639    &  3519  &  3531   &  6  & 83.4\% \\
PQL2H & 1275  & 159   & 160  & 2 & 12.5\%   \\
PQL3H & 1649  & 206   & 207  & 3 & 45.2\% \\
PQL+ & 2924  & 365   & 367  & 3 & 30.6\%   \\
\hline
\end{tabular}
}
\caption{\fontsize{10}{12}\selectfont Statistics of datasets. To count the data percentage with more than one path, \emph{i.e.} \textgreater 1 path, we use graph search algorithm to calculate what percentage of QA pairs can be solved with multiple reasoning paths. %Note that CWQ dataset does not come with relation path annotation. We get max\_hops by counting the number of relations used in the SPARQL language.
}\label{tab:stats}
\end{table}

For questions with multiple answers, we use each answer to construct a question-answer (QA) pair. For WQSP and CWQ, we build a subgraph in a similar way as in \cite{DBLP:conf/emnlp/SunDZMSC18}, in order to generate the entity and relation candidates. For PQL, the original paper provides a subgraph of the Freebase. We implement our model using \textsc{tensorflow-1.11.0} and choose S-MART \cite{DBLP:journals/corr/YangC16a} and AllenNLP \cite{Gardner2017AllenNLP} as our entity linking tools. %\footnote{Entity linking results on \textsc{WebQuestionSP} can be found at \url{https://github.com/scottyih/STAGG}}.
If multiple topic entities are extracted, we use each topic entity to construct a question-answer pair. We test three different graph embedding methods \textsc{Word2vec} \cite{DBLP:journals/corr/abs-1301-3781}, \textsc{TransE} \cite{DBLP:conf/nips/BordesUGWY13}, and \textsc{HolE} \cite{DBLP:journals/corr/TrouillonN17}, and decide to use \textsc{TransE} in our final experiment based on validation performance. The threshold $k_1$ is set to be: 15 plus the number of answers in the ground truth answer set, and $k_2$ is top 50\%. We adopt the average F1 score and the set accuracy as our main evaluation metrics. %and compare our method with the following methods: 
It is worth noticing that: except our methods' results, all other experimental results are obtained from early published papers. Details of these models can be found from our referenced papers.

%The best tuned parameters on development set are summarized as follows: the dynamic function for RNNs is chosen as a gated recurrent units (GRU) with 2 layers and at most 30 units in decoder. The size of the GRU unit and all other embedding layers are set as 300. The threshold $k_1$ is set to be: 15 plus the number of answers in the ground truth answer set, and $k_2$ is top 50\%. We set the dropout rate as 0.3, and train the model using an Adam optimizer with a learning rate of $0.0005$. The Beam size is set to be 12 for both training and inference.  We adopt the average F1 score and the set accuracy as our main evaluation metrics. %and compare our method with the following methods: It is worth noticing that: except our methods' results, all other experimental results are obtained from early published papers. Details of these models can be found from our referenced papers.

%We test different objective functions on WQSP and CWQ datasets. For WQSP, the dataset is originally labeled with multiple relation paths for each QA pair. We thus use them to train our model with joint probability objective. For CWQ, which does not come with any labeled relation path, we parse the given SPARQL queries into relation paths as ground truth labels in single path experiment. The multiple paths experiment is done by using both DFS and trained single path model to extract extra relation paths which points to the answer.

\subsection{Experimental Results}

 In Table \ref{tab:wqsp_cwq} we compare our method to state-of-the-art models. All comparisons are divided into two groups based on different training supervisions. The upper block shows methods that are only trained with final answer as supervision, and the second block contains methods using extra annotations such as parsing results of the query. Experimental results show that our model performs better than all other methods on two datasets except for NSM \cite{DBLP:conf/acl/LiangBLFL17} on WQSP. Although NSM only relies on answers to train their model, it requires many prior knowledges, such as a big vocabulary to train word embeddings and graph embeddings, type label of the entity and of the relation, and pre-defined templates. The experiments from their papers show that these knowledge play a very important role in the system, \emph{e.g.} F1 score drops from 69.0 to 60.7 by not using the pretrained embeddings. %In contrast, our model supports a training method that takes only raw QA pairs and the facts in knowledge base, and does not rely on any additional labels and pre-defined knowledge. 
 Also, NSM is only tested on a single dataset, \emph{i.e.} WQSP. It is unclear whether they could perform consistently well on different datasets. Among all the methods, \textsc{STAGG} performs the best when additional annotation is provided, but we can see a clear drop between \textsc{STAGG\_SP} and \textsc{STAGG\_answer} when such annotation is not available.

%By comparing performance of using different objectives as shown in the second block of the table, we can see that there is a significant improvement by considering multiple relation paths in training. The performance gap between joint objective and marginal objective demonstrates that our proposed marginal objective is a much better way to train a model with multiple relation paths. We do not observe a very different results by using or not using labeled relation paths, which is a good signal. 

\begin{table}[h]\centering
\resizebox{1.01\columnwidth}{!}{
\begin{tabular}{|l|c|c|}
\hline
              & WQSP & CWQ \\
\hline
STAGG\_SP \cite{DBLP:conf/acl/YihRMCS16} & 71.7 & -   \\
HR-BiLSTM \cite{DBLP:conf/acl/YuYHSXZ17}         & 62.3 & 31.2     \\
KBQA-GST \cite{DBLP:conf/ijcai/LanW019}     & 67.9 & 36.5   \\
\hline
KV-MemNN* \cite{DBLP:conf/emnlp/MillerFDKBW16}  & 38.6 & -   \\
STAGG\_answer* \cite{DBLP:conf/acl/YihRMCS16} & 66.8 & -   \\
NSM* \cite{DBLP:conf/acl/LiangBLFL17} & \textbf{69.0} & -   \\
GRAFT-Net* \cite{DBLP:conf/emnlp/SunDZMSC18}        & 62.8 & 26.0   \\

%Our Method-joint\_prob       &   62.1 &  38.0    \\
%Our Method-joint\_prob\_short       &   58.4 &      \\
%Our Method-joint\_prob\_random       &   58.8 &      \\
%Our Method-joint\_prob\_multiple\_paths       & 63.9   &   -    \\
%Our Method-marginal\_prob\_with\_true\_label       &  \textbf{68.5}  &    34.8   \\
Our Method-marginal\_prob*       &  67.9  &   \textbf{41.9}  \\
%Our Method-obj3+new\_decode &   &     \\
\hline
\end{tabular}
}
\caption{\fontsize{10}{12}\selectfont We report F1 ($\%$) on WQSP and CWQ test sets. Methods labeled with $*$ only require the final answer as the supervision, and they are directly comparable to our model. As references, We also report the performance of methods that requires extra supervisions in the first block.}\label{tab:wqsp_cwq}
\end{table}

\begin{table}[h]\centering
\resizebox{0.8\columnwidth}{!}{
\begin{tabular}{|l|c|c|}
\hline
Setting             & $\Delta$ F1 (std) \\
\hline
$\textsc{Ours} - entity\_in\_RNN$     & $-2.1$ (0.21)   \\
$\textsc{Ours} - marginal\_prediction$     & $-1.8$ (0.32)    \\
$\textsc{Ours} - inference\_in\_training$     & $-3.4$ (0.15)   \\
$\textsc{Ours} - mutual\_information$     & $-1.8$ (0.16)   \\
\hline
\end{tabular}
}
\caption{\fontsize{10}{12}\selectfont Feature ablation study on the dev set with a mean of 5 runs.}\label{tab:wqsp_cwq_ablation}
\end{table}

To further disentangle the contribution of different factors in our method, we present a feature ablation test on WQSP dataset shown in Table \ref{tab:wqsp_cwq_ablation}. The vanilla RNN structure only maintains a hidden state and the previous prediction in the loop. Here, we show the performance boost by considering entity features in KBQA task. Instead of using greedy algorithm or beam search to output the top prediction with the highest joint probability $P(y,\mathbf{p})$, we propose to make the prediction based on marginalized probability $P(y)$, which also improves the performance by $1.8\%$. In addition, we show the benefits of using inference during training (line 6 and 7 in algorithm \ref{alg:train}) and mutual information objective (Section (\ref{sec:pmi})). More discussions can be found in the Section \ref{sec:case}. 

\begin{table*}[t]
  \centering
  \begin{tabular}{|c|c|c|}
  \hline
  Method & Objective & Path $\mathbf{p}$ \\
  \hline
     single ground truth& $ p(y|\mathbf{p},q)p(\mathbf{p}|q)$ & single ground truth path leading to $y$\\
 %   \hline
 %   single shortest& $ p(y|\mathbf{p},q)p(\mathbf{p}|q)$ & single shortest path leading to $y$\\
    \hline
    single random& $ p(y|\mathbf{p},q)p(\mathbf{p}|q)$ & single random path leading to $y$\\
     \hline
        multiple product& $\prod_{\mathbf{p}\in\mathcal{P}} p(y|\mathbf{p},q)p(\mathbf{p}|q)$ & all valid paths leading to $y$\\
        \hline
     multiple marginal (ours)& $\sum_{\mathbf{p}\in\mathcal{P}} p(y|\mathbf{p},q)p(\mathbf{p}|q)$ & all valid paths leading to $y$\\
     \hline
  \end{tabular}
  \caption{Different choices of paths and objectives.}
  \label{tab:obj_fcn}
\end{table*}

\begin{table}[h]\centering
\resizebox{1.0\columnwidth}{!}{
\begin{tabular}{|l|c|c|c|c|c|c|}
\hline
                & \multicolumn{3}{c|}{WQSP}   & \multicolumn{3}{c|}{CWQ}   \\ \hline
                & 1 path & \textgreater{}1 path & all & 1 path & \textgreater 1 path & all \\ \hline
single ground truth     & 60.8  & 63.3   & 62.1 &   32.8   &  41.2& 38.4        \\ 
%single shortest & 60.0  & 57.0    &58.4&    30.2    & 40.0    &  36.8         \\ 
single random & 59.7   & 58.1    &58.8&   32.8    & 38.9   & 36.9          \\ 
multiple product &  63.1  &  64.2   &63.7&   32.9    & 42.7    & 39.5          \\ 
multiple marginal (ours)   & 66.0  & 69.3    &67.9&35.7    & 45.0 & 41.9        \\ \hline
\end{tabular}}
\caption{\fontsize{10}{12}\selectfont We break test set into two groups based on number of paths associated with them and report F1($\%$).}\label{tab:wqsp_cwq_path_break}
\end{table}

\subsection{Choices of paths} \label{sec:cop}
In the second set of experiment, we test our model with different objective functions and compare their results correspondingly. The objective functions are as defined in Table \ref{tab:obj_fcn}, where the paths used for training are given in the last column. The detailed explanations are given as following:

\noindent{\bf Single ground truth path.} 
When one reasoning path is given for each QA pair in addition to the answer, we can train the model to fit the given path and answer by maximizing $p(y,\mathbf{p}|q)=p(y|\mathbf{p},q)p(\mathbf{p}|q)$. This objective ignores the fact that multiple reasoning paths could be valid for the same answer (see Figure \ref{QAPaths}) and pushes all the probability mass to the single given one.

\noindent{\bf Single random path.} 
Many existing methods require a ground truth path for each question in order to train the model.
When only the ground truth answer but no path is given to each question, one can randomly sample a path that leads to the given answer and treat the sampled path as ground truth for training.

%\noindent{\bf Single shortest path.} 
%Another way of selecting a path when only the answer entity is given is to choose the shortest path that leads to the given answer entity.

\noindent{\bf Multiple paths product.} 
For many of the existing training methods which expect a single path leading to the answer as part of the input, it is also possible to make them incorporate multiple possible paths when no path annotation is given. The simplest way is to expand each (question, answer) pair into multiple training instances, each with a different path leading to the same answer, and then apply existing training method treating them as independent instances. This corresponds to the objective $\prod_{\mathbf{p}\in\mathcal{P}} p(y|\mathbf{p},q)p(\mathbf{p}|q)$.
This objective has an undesired consequence in practical model training: because of the multiplication operation, the model has to assign equally high probabilities to all given reasoning paths in order to maximize the product of the probabilities. If only some reasoning paths receive high probabilities while others receive low probabilities, the production will still be low. As a consequence, the model cannot differentiate bad reasoning paths from good ones by assigning distinguishable probabilities to them.
 
 \begin{table}[h]\centering
\resizebox{1.01\columnwidth}{!}{%resize the table
\begin{tabular}{|l|c|c|c|}
\hline
              & PQL2H & PQL3H & PQL+ \\
\hline
HR-BiLSTM \cite{DBLP:conf/acl/YuYHSXZ17}         & 97.5 & 87.9 & 92.9    \\
IRN \cite{DBLP:conf/coling/ZhouHZ18}             & 72.5 & 71.0   & 52.9   \\
ABWIM \cite{DBLP:journals/corr/abs-1801-09893}           & 94.3 & 89.3 & 92.6     \\
UHop \cite{DBLP:conf/naacl/ChenCCNK19}            & 97.5 & 89.3 & 92.3   \\
\hline
KV-MemNN* \cite{DBLP:conf/emnlp/MillerFDKBW16}  & 72.2 & 67.4 & -     \\
%Our Method-joint\_prob-nomemory     & 95.3 & 94.8 & 94.5    \\
%Our Method-joint\_prob       & 98.3 & 97.1 & 97.5     \\
Our Method-marginal\_prob*     & \textbf{98.4} & \textbf{97.8} & \textbf{98.0}    \\
\hline
\end{tabular}
}
\caption{\fontsize{10}{12}\selectfont We report set accuracy ($\%$) on PQL. Similar to Table \ref{tab:wqsp_cwq}, we use $*$ to highlight the methods which only requires the answer as supervision.}\label{tab:qpl}
\end{table}
 
\noindent{\bf Multiple paths marginalization.} 
Our proposed training objective replaces the multiplication operation by the summation operation, and this allows the model to concentrate only on good reasoning paths for each QA pair. It is easy to show that the model tends to assign high probability $p(\mathbf{p}|q)$ to a path $\mathbf{p}$ when the path leads to few possible answers and therefore the chance of getting the correct answer $p(y|\mathbf{p},q)$ is high (see ~\ref{eq:equal}). Also, using Jensen's inequality, one can show that this marginal probability objective maximizes the answer probability directly which is the learning goal of KBQA task, while the previous one using product operation maximizes a lower bound. %A benefit of using our model with this objective is that the key hashing process filters out irrelevant memory slots from the full search space. During training, the smaller relation paths that point to an answer set, the larger probability mass will be assigned to this relation path. %This feature satisfies our definition of good path in the introduction. 

We test different ways of choosing paths and defining training objectives on WQSP and CWQ datasets. We further divide the test samples into two groups, based on whether there exist multiple possible paths between the topic entity and the answer based on KB. Table \ref{tab:wqsp_cwq_path_break} show that our proposed method gives the best performance on both scenarios. The models trained with single path perform consistently worse than those trained with multiple paths. Using random ath is worse than using the given ground truth path. Between two models trained using multiple paths, the result shows the advantage of using our proposed objective.

\begin{table*}[t]\centering
\resizebox{2.1\columnwidth}{!}{%resize the table
\begin{tabular}{l|l|l}
\hline
\multicolumn{3}{c}{Question: what state does romney live in? \ \ \    Answer: Massachusetts \ \ \    Topic entity: romney}                                                                      \\ \hline
\textsc{Single\ Ground\ truth} & \textsc{Multiple\ product} & \textsc{Multiple\ marginal (our)}\\ \hline
.89:children & .29:education\_institution/\ state\_province\_region & .83:\textbf{places\_lived/\ location} \\ \hline
.06:\textbf{government\_positions/\ jurisdiction\_of\_office} & .25:\textbf{places\_lived/\ location} & .12:\textbf{government\_positions/\ jurisdiction\_of\_office} \\ \hline
.04:\textbf{government\_positions/\ office\_position\_or\_title} & .25:\textbf{government\_positions/\ district\_represented} & .04:\textbf{government\_positions/\ district\_represented} \\ \hline
.00:\textbf{government\_positions/\ district\_represented} & .01:\textbf{government\_positions/\ jurisdiction\_of\_office} & .01:place\_of\_birth/\ state \\ \hline
.00:place\_of\_birth & .01:place\_of\_birth/\ state & .00:education/\ degree \\ \hline
.00:jurisdiction\_of\_office & .01:sibling/\ place\_of\_birth & .00:election\_campaigns \\ \hline

\end{tabular}
}\caption{\fontsize{10}{12}\selectfont A running examples from \textsc{WebQuestionSP} dataset. We show the probability $P(r_0,\cdots,r_T|q)$ before the inferred relations. Paths that lead to the correct answers are highlighted in bold. We use $/$ to split two relations. The three columns are corresponding to the results by using different training settings as it is in Table \ref{tab:obj_fcn}. Due to space limit, we only show the partial name of a relation in the example and the probability less than .01 is shown as .00. We do not show $P(e_0,\cdots,e_{T-1}|q)$ because they are not determined by our model. }\label{tab:case}
\end{table*}

\subsection{PathQuestion-Large}

In the third set of experiments, we test our model on \textsc{PathQuestion-Large} (PQL) dataset. This dataset contains synthetic questions generated by templates, and is supported by a very small knowledge base (500,000 times smaller than the full freebase). Not surprisingly, we can see the average performance on this dataset is much better than it is on the other two datasets. Recall that PQL2H and PQL3H represents two subsets with only 2 hops and 3 hops questions respectively. Table \ref{tab:qpl} shows that our method's performance beats all the other approaches on all three subsets of PQL from 1$\%$ to 7.8$\%$ in terms of test accuracy. Especially the gap between our method to the previous state-of-the-art approach (\emph{i.e.} UHop) becomes larger when the number of hops increase from 2 to 3.

\section{Case Study}\label{sec:case}

%\subsection{Threshold Analysis}

Our model requires inference while using the current model to select training samples for next batch in training (see line 6 in Algorithm \ref{alg:train}). This EM style training approach helps us filter out bad reasoning paths based on context information. For example, a sample question from WQSP is \textit{who was the owner of kfc?}, the graph search algorithm can easily extract two ``correct'' paths starting from the topic entity \textit{kfc} directing to the ground truth answer \textit{Colonel Sanders}: \textit{kfc $\rightarrow$ organization.organization.founders $\rightarrow$ Colonel Sanders} and \textit{kfc $\rightarrow$ advertisingcharacters.product.advertising\_characters $\rightarrow$ Colonel Sanders}. However, the second path is totally wrong given that the reasoning path is irrelevant to the given question. \textit{Colonel Sanders} happens to be the advertising character of \textit{kfc}, but this cannot be generalized to other cases. Without using the trained model to filter out this irreverent path, the model may learn incorrect map from \textit{who is the owner...} to the relation \textit{advertising\_characters}. In our experiment, we observe that when we train our model with all reasoning paths generated from DFS algorithm without using this filtering strategy (\emph{i.e.} $k_2=inf$), the F1 score drops $3.4\%$ as shown in Table \ref{tab:wqsp_cwq_ablation}. This shows the importance of using the filtering strategy.

Next we demonstrate the benefit of maximizing conditional mutual information instead of likelihood. A sample question in WQSP is \textit{who did benjamin franklin get married to?}. We observe that there are 13 questions are using \textit{Benjamin Franklin} as the topic entity in the training set, but most of them are related to his invention and none of them is about marriage. With such a strong prior on \textit{Benjamin Franklin}, our experimental result shows that the model trained with maximum likelihood mistakingly maps this question to a path related to \textit{invention}, while the model trained with mutual information makes the correct prediction. Table \ref{tab:wqsp_cwq_ablation} shows that we get a $1.8\%$ performance boost by using mutual information.

%\subsection{Case Analysis}

We further show how generated probabilities look like with different choices of paths and objectives in Table \ref{tab:case}. In the given example, only our method outputs the correct path, and one can also find that the top three results correspond to three different but correct reasoning processes. We observe that in many training questions \textit{``live in''} co-occurs with word \textit{``children"}, which explains why the first model makes wrong prediction. %In the second example, the top one ranked path of our model is not correct. However, with marginalizing out  strategy, it re-discover the correct answer by considering cumulative probabilities of it.
We can see that training with joint objective given a single relation path generates the most sharp relation path distribution, \emph{i.e.} the gap between the top entity and the second one is larger than that using other objectives. It assigns most probability mass to the top relation path. In this case, the model does not have ability to identify multiple relation paths during inference. The other extreme is that the second model is trained with joint objective and multiple input paths, which distribute probabilities over many relation paths, hence the model cannot distinguish good relation paths from the bad ones. Between the above two extremes is the proposed marginal objective with multiple input paths, when the most probable path is assigned the largest probability, while the rest ones still get reasonable probability assignments. %In this case, it is possible to apply ensemble prediction method to combine the results of beam search outputs and re-rank them. %In our experiment, we observe 1\% performance boost on WQSP in terms of F1 using ensemble prediction method instead of simply selecting the top searching result of beam search as the prediction.

%Another interesting observation is that given the probability of the predicted relation, our model can judge the correctness of the predictions. By filtering out the test samples where the model do not show confidence (\emph{i.e.} probability of the top relation path is less than 0.9), we observe significant improvements on all three models. For example, the F1 score of the third model improves from 0.685 to 0.774 after removing 489 samples on WQSP. It shows the potential to use a second model to further re-rank our prediction results.

%\subsection{Knowledge based Question Answering (KB-QA)}
\section{Related Work}
%Most of the early KB-QA models are designed for QA problems that can be answered by a single hop, where one hop is defined as the traversal from one entity to another via a single relation. 

Most of the existing multi-hop KBQA systems %\cite{DBLP:conf/acl/YuYHSXZ17,DBLP:journals/corr/abs-1801-09893,DBLP:conf/coling/ZhouHZ18,DBLP:conf/naacl/ChenCCNK19} 
approach this task by decomposing it into two sub-tasks:
%Currently, there are mainly two main categories of methods to tackle the complex KBQA problems, the first category is based on semantic parsing, and the other category mainly relies on the embeddings (XXX) for information retrieval (IR). %Their differences will be discussed further in the relation work section, and i
%In this paper, we focus on the second category, \emph{i.e.} the embedding based approach, which can either predict the answer directly (XXX) or search (a) relation path(s) leading to the final answer entity (XXX). Conventionally, the relation path searching algorithm consists two main subtasks
topic--entity linking and relation extraction. The topic--entity linking gives the system an entry point to start searching, and the relation extraction is used to search relation paths leading to the final answer. %Unlike the simple single relation questions, for complex questions, the path to the final answer may contains multiple hops, where one hop is defined as a searching step between one entity to another via a single relation. 
%For entity linking, there exists many off-the-shell tools  \cite{DBLP:journals/corr/YangC16a} that can give decent performance, and most of the current KBQA models rely on them. For relation extraction, traditional approaches consider all the paths as candidates and search for the best one among them using ranking algorithms. 
%Following this track, existing multi-hop KBQA approaches can be categorized into three groups. 
Following this track, a straightforward idea is to match the question to a candidate entity/relation directly via calculating the similarity between them \cite{DBLP:journals/corr/abs-1801-09893,DBLP:conf/adbis/YuHYZW18,DBLP:conf/ijcai/LanW019}. This method is not ideal for multi-hop questions with long paths, because the number of candidate entity-relation combinations grows exponentially as the number of hops increases. To tackle this issue, methods are proposed to decompose the input question into several single-hop questions, and then use existing method to solve each simple question. The decomposition methods are based on semantic parsing \cite{DBLP:conf/www/AbujabalYRW17,DBLP:conf/emnlp/LuoLLZ18} or templates \cite{DBLP:journals/corr/abs-1908-11053}. A similar idea is to encode the reasoning information hop by hop, and predict the final answer at the last hop \cite{DBLP:conf/emnlp/MillerFDKBW16,DBLP:conf/coling/ZhouHZ18,DBLP:conf/naacl/ChenCCNK19}. %which is a group of methods that are more capable of handling multi-hop questions. 
%\newcite{DBLP:conf/acl/LiYSLYCZL19} casts the task as a multi-turn question answering problem, where the extraction of entities and relations is transformed to the task of identifying answers of template questions. 
%It is also popular to enhance ERL results by using multi-task learning with the help of additional annotations \cite{DBLP:conf/aaai/DengXLYDFLS19,DBLP:conf/ijcai/ShaoGBJCLD19}. 

Another line of work has looked at solving KBQA task with only final answer as supervision. \newcite{DBLP:conf/acl/LiangBLFL17} first propose to cast KBQA as a program generation task using neural program induction (NPI) techniques. They learn to translate the query to a program like logical form executable on the KB. As a follow up, \newcite{DBLP:conf/ijcai/AnsariSKBSC19} improves this idea by incorporating high level program structures. Both these NPI models do not require annotated relation path as supervision, but they need some prior knowledge to design the program templates. In other work, \newcite{DBLP:journals/corr/abs-1909-04849} recently proposed a latent variable approach which is similar to the one described here, but applied on text-based QA scenarios. The main difference between our work is that our method aims at finding multiple reasoning paths leading to the answer, while their method only focus on extracting single optimal solution. We employ inference during training to filter our irrelevant paths, while they use it to identify the optimal solution.
\section{Conclusion}

In this paper, We introduce a novel KBQA system which can leverage information from multiple reasoning paths. To train our model, we use a marginalized probability objective function. Experimental results show that our model achieve strong performance on popular KBQA datasets.

\clearpage
\newpage

\bibliography{ref}
\bibliographystyle{acl_natbib}
\end{document}